\documentclass[letterpaper,compsoc,twoside]{IEEEtran}
\usepackage{fixltx2e} 
\usepackage{cmap} 
\usepackage{ifthen}
\usepackage[T1]{fontenc}
\usepackage[utf8]{inputenc}
\usepackage{amsmath}

\usepackage[font={small,it},labelfont=bf]{caption}
\usepackage{float}

\pdfoutput=1
\usepackage{scipy}
\makeatletter
\def\PY@reset{\let\PY@it=\relax \let\PY@bf=\relax%
    \let\PY@ul=\relax \let\PY@tc=\relax%
    \let\PY@bc=\relax \let\PY@ff=\relax}
\def\PY@tok#1{\csname PY@tok@#1\endcsname}
\def\PY@toks#1+{\ifx\relax#1\empty\else%
    \PY@tok{#1}\expandafter\PY@toks\fi}
\def\PY@do#1{\PY@bc{\PY@tc{\PY@ul{%
    \PY@it{\PY@bf{\PY@ff{#1}}}}}}}
\def\PY#1#2{\PY@reset\PY@toks#1+\relax+\PY@do{#2}}

\expandafter\def\csname PY@tok@gd\endcsname{\def\PY@tc##1{\textcolor[rgb]{0.63,0.00,0.00}{##1}}}
\expandafter\def\csname PY@tok@gu\endcsname{\let\PY@bf=\textbf\def\PY@tc##1{\textcolor[rgb]{0.50,0.00,0.50}{##1}}}
\expandafter\def\csname PY@tok@gt\endcsname{\def\PY@tc##1{\textcolor[rgb]{0.00,0.27,0.87}{##1}}}
\expandafter\def\csname PY@tok@gs\endcsname{\let\PY@bf=\textbf}
\expandafter\def\csname PY@tok@gr\endcsname{\def\PY@tc##1{\textcolor[rgb]{1.00,0.00,0.00}{##1}}}
\expandafter\def\csname PY@tok@cm\endcsname{\let\PY@it=\textit\def\PY@tc##1{\textcolor[rgb]{0.25,0.50,0.56}{##1}}}
\expandafter\def\csname PY@tok@vg\endcsname{\def\PY@tc##1{\textcolor[rgb]{0.73,0.38,0.84}{##1}}}
\expandafter\def\csname PY@tok@m\endcsname{\def\PY@tc##1{\textcolor[rgb]{0.13,0.50,0.31}{##1}}}
\expandafter\def\csname PY@tok@mh\endcsname{\def\PY@tc##1{\textcolor[rgb]{0.13,0.50,0.31}{##1}}}
\expandafter\def\csname PY@tok@cs\endcsname{\def\PY@tc##1{\textcolor[rgb]{0.25,0.50,0.56}{##1}}\def\PY@bc##1{\setlength{\fboxsep}{0pt}\colorbox[rgb]{1.00,0.94,0.94}{\strut ##1}}}
\expandafter\def\csname PY@tok@ge\endcsname{\let\PY@it=\textit}
\expandafter\def\csname PY@tok@vc\endcsname{\def\PY@tc##1{\textcolor[rgb]{0.73,0.38,0.84}{##1}}}
\expandafter\def\csname PY@tok@il\endcsname{\def\PY@tc##1{\textcolor[rgb]{0.13,0.50,0.31}{##1}}}
\expandafter\def\csname PY@tok@go\endcsname{\def\PY@tc##1{\textcolor[rgb]{0.20,0.20,0.20}{##1}}}
\expandafter\def\csname PY@tok@cp\endcsname{\def\PY@tc##1{\textcolor[rgb]{0.00,0.44,0.13}{##1}}}
\expandafter\def\csname PY@tok@gi\endcsname{\def\PY@tc##1{\textcolor[rgb]{0.00,0.63,0.00}{##1}}}
\expandafter\def\csname PY@tok@gh\endcsname{\let\PY@bf=\textbf\def\PY@tc##1{\textcolor[rgb]{0.00,0.00,0.50}{##1}}}
\expandafter\def\csname PY@tok@ni\endcsname{\let\PY@bf=\textbf\def\PY@tc##1{\textcolor[rgb]{0.84,0.33,0.22}{##1}}}
\expandafter\def\csname PY@tok@nl\endcsname{\let\PY@bf=\textbf\def\PY@tc##1{\textcolor[rgb]{0.00,0.13,0.44}{##1}}}
\expandafter\def\csname PY@tok@nn\endcsname{\let\PY@bf=\textbf\def\PY@tc##1{\textcolor[rgb]{0.05,0.52,0.71}{##1}}}
\expandafter\def\csname PY@tok@no\endcsname{\def\PY@tc##1{\textcolor[rgb]{0.38,0.68,0.84}{##1}}}
\expandafter\def\csname PY@tok@na\endcsname{\def\PY@tc##1{\textcolor[rgb]{0.25,0.44,0.63}{##1}}}
\expandafter\def\csname PY@tok@nb\endcsname{\def\PY@tc##1{\textcolor[rgb]{0.00,0.44,0.13}{##1}}}
\expandafter\def\csname PY@tok@nc\endcsname{\let\PY@bf=\textbf\def\PY@tc##1{\textcolor[rgb]{0.05,0.52,0.71}{##1}}}
\expandafter\def\csname PY@tok@nd\endcsname{\let\PY@bf=\textbf\def\PY@tc##1{\textcolor[rgb]{0.33,0.33,0.33}{##1}}}
\expandafter\def\csname PY@tok@ne\endcsname{\def\PY@tc##1{\textcolor[rgb]{0.00,0.44,0.13}{##1}}}
\expandafter\def\csname PY@tok@nf\endcsname{\def\PY@tc##1{\textcolor[rgb]{0.02,0.16,0.49}{##1}}}
\expandafter\def\csname PY@tok@si\endcsname{\let\PY@it=\textit\def\PY@tc##1{\textcolor[rgb]{0.44,0.63,0.82}{##1}}}
\expandafter\def\csname PY@tok@s2\endcsname{\def\PY@tc##1{\textcolor[rgb]{0.25,0.44,0.63}{##1}}}
\expandafter\def\csname PY@tok@vi\endcsname{\def\PY@tc##1{\textcolor[rgb]{0.73,0.38,0.84}{##1}}}
\expandafter\def\csname PY@tok@nt\endcsname{\let\PY@bf=\textbf\def\PY@tc##1{\textcolor[rgb]{0.02,0.16,0.45}{##1}}}
\expandafter\def\csname PY@tok@nv\endcsname{\def\PY@tc##1{\textcolor[rgb]{0.73,0.38,0.84}{##1}}}
\expandafter\def\csname PY@tok@s1\endcsname{\def\PY@tc##1{\textcolor[rgb]{0.25,0.44,0.63}{##1}}}
\expandafter\def\csname PY@tok@gp\endcsname{\let\PY@bf=\textbf\def\PY@tc##1{\textcolor[rgb]{0.78,0.36,0.04}{##1}}}
\expandafter\def\csname PY@tok@sh\endcsname{\def\PY@tc##1{\textcolor[rgb]{0.25,0.44,0.63}{##1}}}
\expandafter\def\csname PY@tok@ow\endcsname{\let\PY@bf=\textbf\def\PY@tc##1{\textcolor[rgb]{0.00,0.44,0.13}{##1}}}
\expandafter\def\csname PY@tok@sx\endcsname{\def\PY@tc##1{\textcolor[rgb]{0.78,0.36,0.04}{##1}}}
\expandafter\def\csname PY@tok@bp\endcsname{\def\PY@tc##1{\textcolor[rgb]{0.00,0.44,0.13}{##1}}}
\expandafter\def\csname PY@tok@c1\endcsname{\let\PY@it=\textit\def\PY@tc##1{\textcolor[rgb]{0.25,0.50,0.56}{##1}}}
\expandafter\def\csname PY@tok@kc\endcsname{\let\PY@bf=\textbf\def\PY@tc##1{\textcolor[rgb]{0.00,0.44,0.13}{##1}}}
\expandafter\def\csname PY@tok@c\endcsname{\let\PY@it=\textit\def\PY@tc##1{\textcolor[rgb]{0.25,0.50,0.56}{##1}}}
\expandafter\def\csname PY@tok@mf\endcsname{\def\PY@tc##1{\textcolor[rgb]{0.13,0.50,0.31}{##1}}}
\expandafter\def\csname PY@tok@err\endcsname{\def\PY@bc##1{\setlength{\fboxsep}{0pt}\fcolorbox[rgb]{1.00,0.00,0.00}{1,1,1}{\strut ##1}}}
\expandafter\def\csname PY@tok@kd\endcsname{\let\PY@bf=\textbf\def\PY@tc##1{\textcolor[rgb]{0.00,0.44,0.13}{##1}}}
\expandafter\def\csname PY@tok@ss\endcsname{\def\PY@tc##1{\textcolor[rgb]{0.32,0.47,0.09}{##1}}}
\expandafter\def\csname PY@tok@sr\endcsname{\def\PY@tc##1{\textcolor[rgb]{0.14,0.33,0.53}{##1}}}
\expandafter\def\csname PY@tok@mo\endcsname{\def\PY@tc##1{\textcolor[rgb]{0.13,0.50,0.31}{##1}}}
\expandafter\def\csname PY@tok@mi\endcsname{\def\PY@tc##1{\textcolor[rgb]{0.13,0.50,0.31}{##1}}}
\expandafter\def\csname PY@tok@kn\endcsname{\let\PY@bf=\textbf\def\PY@tc##1{\textcolor[rgb]{0.00,0.44,0.13}{##1}}}
\expandafter\def\csname PY@tok@o\endcsname{\def\PY@tc##1{\textcolor[rgb]{0.40,0.40,0.40}{##1}}}
\expandafter\def\csname PY@tok@kr\endcsname{\let\PY@bf=\textbf\def\PY@tc##1{\textcolor[rgb]{0.00,0.44,0.13}{##1}}}
\expandafter\def\csname PY@tok@s\endcsname{\def\PY@tc##1{\textcolor[rgb]{0.25,0.44,0.63}{##1}}}
\expandafter\def\csname PY@tok@kp\endcsname{\def\PY@tc##1{\textcolor[rgb]{0.00,0.44,0.13}{##1}}}
\expandafter\def\csname PY@tok@w\endcsname{\def\PY@tc##1{\textcolor[rgb]{0.73,0.73,0.73}{##1}}}
\expandafter\def\csname PY@tok@kt\endcsname{\def\PY@tc##1{\textcolor[rgb]{0.56,0.13,0.00}{##1}}}
\expandafter\def\csname PY@tok@sc\endcsname{\def\PY@tc##1{\textcolor[rgb]{0.25,0.44,0.63}{##1}}}
\expandafter\def\csname PY@tok@sb\endcsname{\def\PY@tc##1{\textcolor[rgb]{0.25,0.44,0.63}{##1}}}
\expandafter\def\csname PY@tok@k\endcsname{\let\PY@bf=\textbf\def\PY@tc##1{\textcolor[rgb]{0.00,0.44,0.13}{##1}}}
\expandafter\def\csname PY@tok@se\endcsname{\let\PY@bf=\textbf\def\PY@tc##1{\textcolor[rgb]{0.25,0.44,0.63}{##1}}}
\expandafter\def\csname PY@tok@sd\endcsname{\let\PY@it=\textit\def\PY@tc##1{\textcolor[rgb]{0.25,0.44,0.63}{##1}}}

\def\PYZbs{\char`\\}
\def\PYZus{\char`\_}

\def\PYZgt{\char`\>}
\def\PYZsh{\char`\#}

\def\PYZhy{\char`\-}
\def\PYZsq{\char`\'}

\makeatother

\providecommand*{\DUprovidelength}[2]{
  \ifthenelse{\isundefined{#1}}{\newlength{#1}\setlength{#1}{#2}}{}
}

\providecommand*{\DUrole}[2]{%
  \ifcsname DUrole#1\endcsname%
    \csname DUrole#1\endcsname{#2}%
  \else
    \ifcsname docutilsrole#1\endcsname%
      \csname docutilsrole#1\endcsname{#2}%
    \else%
      #2%
    \fi%
  \fi%
}

\DUprovidelength{\DUlineblockindent}{2.5em}
\ifthenelse{\isundefined{\DUlineblock}}{
  \newenvironment{DUlineblock}[1]{%
    \list{}{\setlength{\partopsep}{\parskip}
            \addtolength{\partopsep}{\baselineskip}
            \setlength{\topsep}{0pt}
            \setlength{\itemsep}{0.15\baselineskip}
            \setlength{\parsep}{0pt}
            \setlength{\leftmargin}{#1}}
    \raggedright
  }
  {\endlist}
}{}

\ifthenelse{\isundefined{\hypersetup}}{
  \usepackage[colorlinks=true,linkcolor=blue,urlcolor=blue]{hyperref}
  \urlstyle{same} 
}{}

\begin{document}
\newcounter{footnotecounter}\title{Py3DFreeHandUS: a library for voxel-array reconstruction using Ultrasonography and attitude sensors}\author{Davide Monari$^{\setcounter{footnotecounter}{1}\fnsymbol{footnotecounter}\setcounter{footnotecounter}{2}\fnsymbol{footnotecounter}}$%
          \setcounter{footnotecounter}{1}\thanks{\fnsymbol{footnotecounter} %
          Corresponding author: \protect\href{mailto:davide.monari@kuleuven.be}{davide.monari@kuleuven.be}}\setcounter{footnotecounter}{2}\thanks{\fnsymbol{footnotecounter} KULeuven}, Francesco Cenni$^{\setcounter{footnotecounter}{2}\fnsymbol{footnotecounter}}$, Erwin Aertbeliën$^{\setcounter{footnotecounter}{2}\fnsymbol{footnotecounter}}$, Kaat Desloovere$^{\setcounter{footnotecounter}{2}\fnsymbol{footnotecounter}}$\thanks{%

          \noindent%
          Copyright\,\copyright\,2014 Davide Monari et al. This is an open-access article distributed under the terms of the Creative Commons Attribution License, which permits unrestricted use, distribution, and reproduction in any medium, provided the original author and source are credited. http://creativecommons.org/licenses/by/3.0/%
        }}\maketitle
          \renewcommand{\leftmark}{PROC. OF THE 7th EUR. CONF. ON PYTHON IN SCIENCE (EUROSCIPY 2014)}
          \renewcommand{\rightmark}{PY3DFREEHANDUS: A LIBRARY FOR VOXEL-ARRAY RECONSTRUCTION USING ULTRASONOGRAPHY AND ATTITUDE SENSORS}

\setcounter{page}{43}
\newcommand*{\docutilsroleref}{\ref}
\newcommand*{\docutilsrolelabel}{\label}
\AtEndDocument{\cleardoublepage}
\begin{abstract}In medical imaging, there is a growing interest to provide real-time images with good quality for large anatomical structures. To cope with this issue, we developed a library that allows to replace, for some specific clinical applications, more robust systems such as Computer Tomography (CT) and Magnetic Resonance Imaging (MRI). Our python library \emph{Py3DFreeHandUS} is a package for processing data acquired simultaneously by ultra-sonographic systems (US) and marker-based optoelectronic systems. In particular, US data enables to visualize subcutaneous body structures, whereas the optoelectronic system is able to collect the 3D position in space for reflective objects, that are called markers. By combining these two measurement devices, it is possible to reconstruct the real 3D morphology of body structures such as muscles, for relevant clinical implications. In the present research work, the different steps which allow to obtain a relevant 3D data set as well as the procedures for calibrating the systems and for determining the quality of the reconstruction.\end{abstract}

\section{Introduction%
  \label{introduction}%
}

In medical imaging, 3D data sets are an essential tool to explore anatomical volumes and to extract clinical features, which can describe a particular condition of the patient. These data are usually recorded by CT or MRI for identifying hard or soft tissue, respectively, and provide a high image quality together with a large field of view. On the other hand, these systems are \emph{very} expensive (espacially MRI ones) and time consuming both for operators and patients. Plus, radioations from CT are an issue. Therefore, for some clinical applications, it could be interesting to replace these systems with others that can allow to provide 3D data sets quickly, although without the same high image quality.
Ultrasonography (US) devices are systems largely used to collect medical images. For example, it is very common to examine pregnant women. This system, compared to other medical imaging systems, has several advantages: real-time images, portability, no ionizing radiation. However, one of the major drawbacks in US is the limited field of view and the lack of spatial information among different images acquired. Therefore a technique called 3D Freehand Ultrasound (3DUS) was originally proposed in the 90s \cite{Rankin93}, \cite{Prager99} with the aim of reconstructing large 3D anatomical parts. The idea is to combine US images and the corresponding position and orientation (POS) of the US transducer; by simultaneously scanning a series of 2D images and recording spatial information it is possible to perform the relevant reconstruction and then the visualization of the entire volume acquired.
The aim of the present work is to customize the 3DUS implementation by pushing on vectorization in NumPy / SciPy along with memory waste avoidance, for speeding up the processing phase as much as possible. These aspects are essential in this context, since for commodity hardware: i) memory resources are relatively limited and 3D volumes involved here can quickly reach large dimensions, ii) computation time can become unrealistic if very large for- or while- loops are used in Python. In addition the few existing applications for applying this technique have at least one of the following disadvantages: i) not open-source; ii) only supporting data streams from a limited number of US/POS sensors; iii) they are written in low-level languages such as C++, making rapid development and prototyping more difficult.
We developed a pure Python library called \textbf{Py3DFreeHandUS} that solves all the above issues.

\section{Requirements%
  \label{requirements}%
}

Py3DFreeHandUS was developed in Python 2.7 (Python 3 not yet supported), and uses the following libraries:%
\begin{itemize}

\item 

NumPy
\item 

SciPy (0.11.0+)
\item 

matplotlib
\item 

SymPy
\item 

pydicom
\item 

b-tk (Biomechanical ToolKit) \cite{Barre14}
\item 

VTK
\item 

OpenCV (2.4.9+)
\item 

Cython + gcc (optional, we are \textquotedbl{}cythonizing\textquotedbl{} bottlenecks but leaving pure Python implementation available)
\end{itemize}

We used the Python distribution \emph{Python(x,y)} for development and testing, since it already includes all libraries but b-tk.

\section{Description of the package%
  \label{description-of-the-package}%
}

The present package is able to process synchronized data by US and POS, being as input DICOM and C3D files, respectively. The operations flowchart is composed by: US probe \emph{temporal} and \emph{spatial calibration} and \emph{3D voxel array reconstruction}.

\subsection{US probe temporal calibration%
  \label{us-probe-temporal-calibration}%
}

The aim of the temporal calibration is to estimate the time delay between US and the POS devices. This procedure is foundamental whenever it is not possible to hardware-trigger US and POS devices, so data needs to be time-shifted later. Time delay resolution cannot be lower than the inverse of the lower frequency (normally US). Briefly, we moved vertically up and down the US probe (rigidly connected to the POS sensor) in a water-filled tank and generated two curves: the first one being the vertical coordinate of the the POS sensor, the second one being the vertical coordinate (in the US image) of the center of the line representing the edge between water and tank. These two sine-like signal were demeaned, normalized to be inside the range {[}-1,+1{]} and cross-correlated with the function \texttt{matplotlib.pyplot.xcorr}. The time of the first peak for the cross-correlation estimates the time delay.

\subsection{US probe spatial calibration%
  \label{us-probe-spatial-calibration}%
}

The probe spatial calibration is an essential procedure for image reconstruction which allows to determine the \emph{pose} (position and orientation) of the US images with respect to the POS device. The corresponding results take the form of six parameters, three for position and three for orientation. The quality of this step mainly influences the reconstruction quality of the anatomical shape. To perform the probe calibration we used two different steps. First we applied an established procedure already published in the literature \cite{Prager98} and later we tuned the results by using an image compounding algorithm \cite{Wein08}. The established procedure was proposed by Prager et al. \cite{Prager98} and improved by Hsu \cite{Hsu06}, with the idea of scanning the floor of a water tank by covering all the degrees of freedom (see Figure \DUrole{ref}{calib}); this scanning modality produces clear and consistent edge lines (between water and tank bottom) in the US images (B-scans). All the pixels lying on the visible line in the B-scan should satisfy equations that come from the different spatial transformations, which leave to solve 11 identifiable parameters. Each B-scan can be used to write 2 equations. The overdetermined set of equations is solved using the Levenberg-Marquardt algorithm. We found that it is essential to move the US transducers following the sequence of movements suggested in \cite{Prager98}, in order to have reasonable results. The equation that a pixel with image coordinates $(u,v)$ must satisfy (see \cite{Prager98} for details) is as follows:
\begin{DUlineblock}{0em}
\item[] 
\end{DUlineblock}

$\begin{pmatrix} 0 \\ 0 \\ 0 \\ 1 \end{pmatrix} =\ ^{C}T_{T}\ ^{T}T_{R}\ ^{R}T_{P} \begin{pmatrix} s_{x}u \\ s_{y}v \\ 0 \\ 1 \end{pmatrix}$,
\begin{DUlineblock}{0em}
\item[] 
\end{DUlineblock}

where $s_{x}$ and $s_{y}$ are conversion factors from \emph{pixel} to \emph{mm}.\begin{figure}[]\noindent\makebox[\columnwidth][c]{\includegraphics[width=\columnwidth]{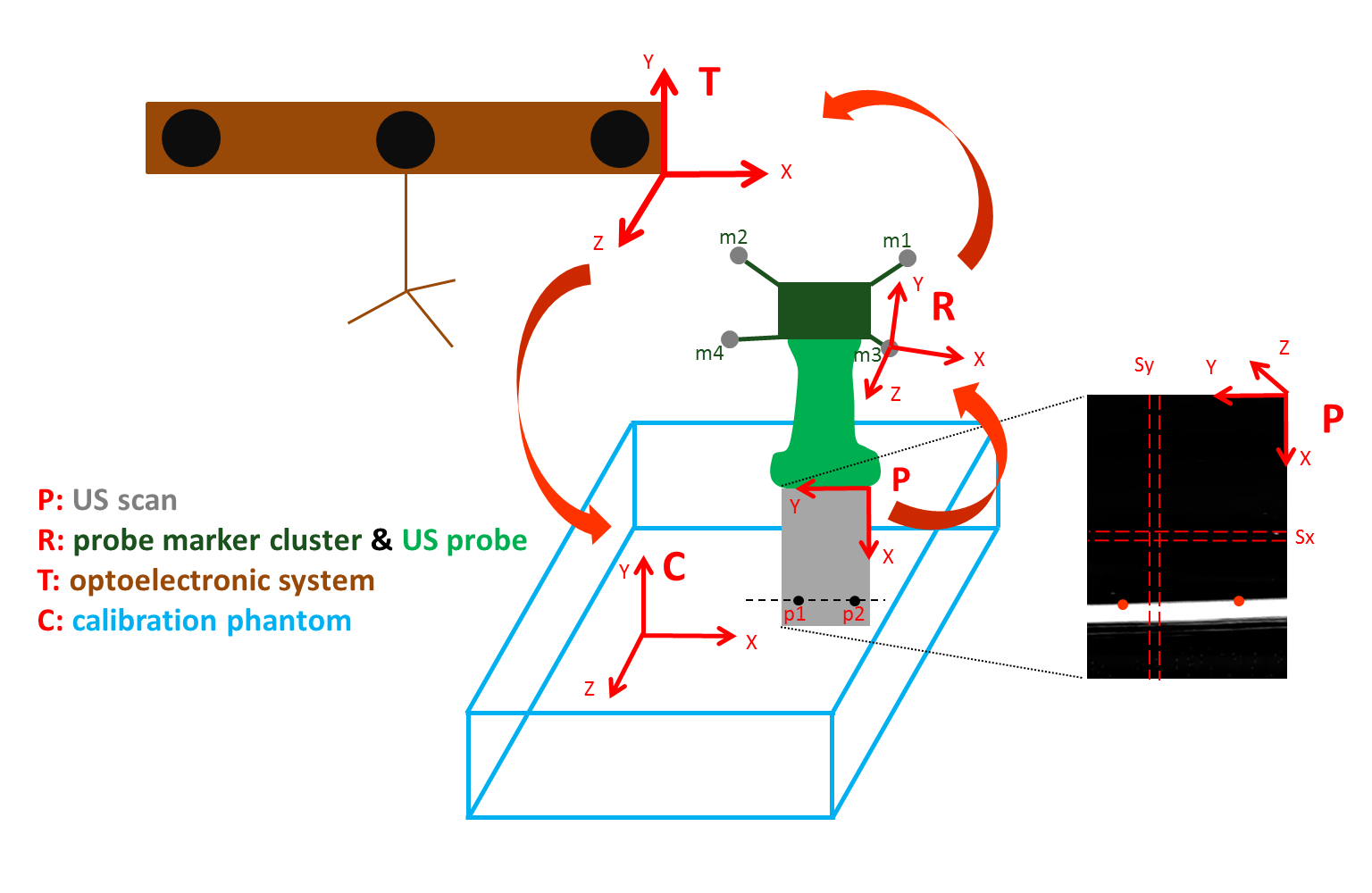}}
\caption{The aim of the US probe spatial calibration is to find the roto-translation matrix $^{R}T_{P}$ from the image reference frame (P) to the transducer reference frame (R). The other two roto-translation matrices $^{T}T_{R}$ and $^{C}T_{T}$ (respectively, from transducer to optoelectronic system and from optoelectronic system to calibration phantom) are known for every time frame of the calibration acquisition. \DUrole{label}{calib}}
\end{figure}

This is the code snippet for the equation creation:
\begin{DUlineblock}{0em}
\item[] 
\end{DUlineblock}
\begin{Verbatim}[commandchars=\\\{\},fontsize=\footnotesize]
\PY{k+kn}{from} \PY{n+nn}{sympy} \PY{k+kn}{import} \PY{n}{Matrix}\PY{p}{,} \PY{n}{Symbol}\PY{p}{,} \PY{n}{var}
\PY{k+kn}{from} \PY{n+nn}{sympy} \PY{k+kn}{import} \PY{n}{cos} \PY{k}{as} \PY{n}{c}\PY{p}{,} \PY{n}{sin} \PY{k}{as} \PY{n}{s}

\PY{c}{\PYZsh{} Pp}
\PY{n}{sx} \PY{o}{=} \PY{n}{Symbol}\PY{p}{(}\PY{l+s}{\PYZsq{}}\PY{l+s}{sx}\PY{l+s}{\PYZsq{}}\PY{p}{)}
\PY{n}{sy} \PY{o}{=} \PY{n}{Symbol}\PY{p}{(}\PY{l+s}{\PYZsq{}}\PY{l+s}{sy}\PY{l+s}{\PYZsq{}}\PY{p}{)}
\PY{n}{u} \PY{o}{=} \PY{n}{Symbol}\PY{p}{(}\PY{l+s}{\PYZsq{}}\PY{l+s}{u}\PY{l+s}{\PYZsq{}}\PY{p}{)}
\PY{n}{v} \PY{o}{=} \PY{n}{Symbol}\PY{p}{(}\PY{l+s}{\PYZsq{}}\PY{l+s}{v}\PY{l+s}{\PYZsq{}}\PY{p}{)}
\PY{n}{Pp} \PY{o}{=} \PY{n}{Matrix}\PY{p}{(}\PY{p}{(}\PY{p}{[}\PY{n}{sx} \PY{o}{*} \PY{n}{u}\PY{p}{]}\PY{p}{,}\PYZbs{}
             \PY{p}{[}\PY{n}{sy} \PY{o}{*} \PY{n}{v}\PY{p}{]}\PY{p}{,}\PYZbs{}
             \PY{p}{[}\PY{l+m+mi}{0}\PY{p}{]}\PY{p}{,}\PYZbs{}
             \PY{p}{[}\PY{l+m+mi}{1}\PY{p}{]}\PYZbs{}
\PY{p}{)}\PY{p}{)}

\PY{c}{\PYZsh{} rTp}
\PY{n}{rTp}\PY{p}{,} \PY{n}{syms} \PY{o}{=} \PY{n}{creatCalibMatrix}\PY{p}{(}\PY{p}{)}
\PY{p}{[}\PY{n}{x1}\PY{p}{,} \PY{n}{y1}\PY{p}{,} \PY{n}{z1}\PY{p}{,} \PY{n}{alpha1}\PY{p}{,} \PY{n}{beta1}\PY{p}{,} \PY{n}{gamma1}\PY{p}{]} \PY{o}{=} \PY{n}{syms}

\PY{c}{\PYZsh{} tTr}
\PY{n}{tTr} \PY{o}{=} \PY{n}{MatrixOfMatrixSymbol}\PY{p}{(}\PY{l+s}{\PYZsq{}}\PY{l+s}{tTr}\PY{l+s}{\PYZsq{}}\PY{p}{,} \PY{l+m+mi}{4}\PY{p}{,} \PY{l+m+mi}{4}\PY{p}{)}
\PY{n}{tTr}\PY{p}{[}\PY{l+m+mi}{3}\PY{p}{,} \PY{l+m+mi}{0}\PY{p}{:}\PY{l+m+mi}{4}\PY{p}{]} \PY{o}{=} \PY{n}{np}\PY{o}{.}\PY{n}{array}\PY{p}{(}\PY{p}{[}\PY{l+m+mi}{0}\PY{p}{,}\PY{l+m+mi}{0}\PY{p}{,}\PY{l+m+mi}{0}\PY{p}{,}\PY{l+m+mi}{1}\PY{p}{]}\PY{p}{)}

\PY{c}{\PYZsh{} cTt}
\PY{n}{x2} \PY{o}{=} \PY{n}{Symbol}\PY{p}{(}\PY{l+s}{\PYZsq{}}\PY{l+s}{x2}\PY{l+s}{\PYZsq{}}\PY{p}{)}
\PY{n}{y2} \PY{o}{=} \PY{n}{Symbol}\PY{p}{(}\PY{l+s}{\PYZsq{}}\PY{l+s}{y2}\PY{l+s}{\PYZsq{}}\PY{p}{)}
\PY{n}{z2} \PY{o}{=} \PY{n}{Symbol}\PY{p}{(}\PY{l+s}{\PYZsq{}}\PY{l+s}{z2}\PY{l+s}{\PYZsq{}}\PY{p}{)}
\PY{n}{alpha2} \PY{o}{=} \PY{n}{Symbol}\PY{p}{(}\PY{l+s}{\PYZsq{}}\PY{l+s}{alpha2}\PY{l+s}{\PYZsq{}}\PY{p}{)}
\PY{n}{beta2} \PY{o}{=} \PY{n}{Symbol}\PY{p}{(}\PY{l+s}{\PYZsq{}}\PY{l+s}{beta2}\PY{l+s}{\PYZsq{}}\PY{p}{)}
\PY{n}{gamma2} \PY{o}{=} \PY{n}{Symbol}\PY{p}{(}\PY{l+s}{\PYZsq{}}\PY{l+s}{gamma2}\PY{l+s}{\PYZsq{}}\PY{p}{)}

\PY{n}{cTt} \PY{o}{=} \PY{n}{Matrix}\PY{p}{(}\PY{p}{(}\PY{p}{[}\PY{n}{c}\PY{p}{(}\PY{n}{alpha2}\PY{p}{)}\PY{o}{*}\PY{n}{c}\PY{p}{(}\PY{n}{beta2}\PY{p}{)}\PY{p}{,} \PY{o}{.}\PY{o}{.}\PY{o}{.}
               \PY{p}{[}\PY{n}{s}\PY{p}{(}\PY{n}{alpha2}\PY{p}{)}\PY{o}{*}\PY{n}{c}\PY{p}{(}\PY{n}{beta2}\PY{p}{)}\PY{p}{,} \PY{o}{.}\PY{o}{.}\PY{o}{.}
               \PY{p}{[}\PY{o}{\PYZhy{}}\PY{n}{s}\PY{p}{(}\PY{n}{beta2}\PY{p}{)}\PY{p}{,} \PY{n}{c}\PY{p}{(}\PY{n}{beta2}\PY{p}{)}\PY{o}{*}\PY{n}{s}\PY{p}{(}\PY{n}{gamma2}\PY{p}{)}\PY{p}{,} \PY{o}{.}\PY{o}{.}\PY{o}{.}
               \PY{p}{[}\PY{l+m+mi}{0}\PY{p}{,} \PY{l+m+mi}{0}\PY{p}{,} \PY{l+m+mi}{0}\PY{p}{,} \PY{l+m+mi}{1}\PY{p}{]}\PYZbs{}
\PY{p}{)}\PY{p}{)} \PY{c}{\PYZsh{} see [Prager98] for full expressions}

\PY{c}{\PYZsh{} Calculate full equations}
\PY{n}{Pc} \PY{o}{=} \PY{n}{cTt} \PY{o}{*} \PY{n}{tTr} \PY{o}{*} \PY{n}{rTp} \PY{o}{*} \PY{n}{Pp}
\PY{n}{Pc} \PY{o}{=} \PY{n}{Pc}\PY{p}{[}\PY{l+m+mi}{0}\PY{p}{:}\PY{l+m+mi}{3}\PY{p}{,}\PY{p}{:}\PY{p}{]}

\PY{c}{\PYZsh{} Calculate full Jacobians}
\PY{n}{x} \PY{o}{=} \PY{n}{Matrix}\PY{p}{(}\PY{p}{[}\PY{n}{sx}\PY{p}{,} \PY{n}{sy}\PY{p}{,} \PY{n}{x1}\PY{p}{,} \PY{n}{y1}\PY{p}{,} \PY{n}{z1}\PY{p}{,} \PY{n}{alpha1}\PY{p}{,} \PY{n}{beta1}\PY{p}{,}
\PY{n}{gamma1}\PY{p}{,} \PY{n}{x2}\PY{p}{,} \PY{n}{y2}\PY{p}{,} \PY{n}{z2}\PY{p}{,} \PY{n}{alpha2}\PY{p}{,} \PY{n}{beta2}\PY{p}{,} \PY{n}{gamma2}\PY{p}{]}\PY{p}{)}
\PY{n}{J} \PY{o}{=} \PY{n}{Pc}\PY{o}{.}\PY{n}{jacobian}\PY{p}{(}\PY{n}{x}\PY{p}{)}
\end{Verbatim}
The equations system was solved by using the function \texttt{scipy.optimize.root} with \texttt{method='lm'}.

To validate the solution, the calibration part in this package allows to visualize the corresponding covariance matrix; this can be exploited to understand if some variable is not well constrained. In addition, since in each B-scan it is necessary to have the position for at least two pixels that belong to the edge line, we developed an automatic tool for extracting the corresponding lines in each image, based on the Hough transform:
\begin{DUlineblock}{0em}
\item[] 
\end{DUlineblock}
\begin{Verbatim}[commandchars=\\\{\},fontsize=\footnotesize]
\PY{k+kn}{import} \PY{n+nn}{cv2}

\PY{c}{\PYZsh{} Threshold image}
\PY{n}{maxVal} \PY{o}{=} \PY{n}{np}\PY{o}{.}\PY{n}{iinfo}\PY{p}{(}\PY{n}{I}\PY{o}{.}\PY{n}{dtype}\PY{p}{)}\PY{o}{.}\PY{n}{max}
\PY{n}{th}\PY{p}{,} \PY{n}{bw} \PY{o}{=} \PY{n}{cv2}\PY{o}{.}\PY{n}{threshold}\PY{p}{(}\PY{n}{I}\PY{p}{,}\PY{n}{np}\PY{o}{.}\PY{n}{round}\PY{p}{(}\PY{n}{thI}\PY{o}{*}\PY{n}{maxVal}\PY{p}{)}\PY{p}{,}
    \PY{n}{maxVal}\PY{p}{,}\PY{n}{cv2}\PY{o}{.}\PY{n}{THRESH\PYZus{}BINARY}\PY{p}{)}
\PY{c}{\PYZsh{} Detect edges}
\PY{n}{edges} \PY{o}{=} \PY{n}{cv2}\PY{o}{.}\PY{n}{Canny}\PY{p}{(}\PY{n}{bw}\PY{p}{,}\PY{n}{thCan1}\PY{p}{,}\PY{n}{thCan2}\PY{p}{,}
    \PY{n}{apertureSize}\PY{o}{=}\PY{n}{kerSizeCan}\PY{p}{)}
\PY{c}{\PYZsh{} Dilate edges}
\PY{n}{kernel} \PY{o}{=} \PY{n}{np}\PY{o}{.}\PY{n}{ones}\PY{p}{(}\PY{n}{kerSizeDil}\PY{p}{,}\PY{n}{I}\PY{o}{.}\PY{n}{dtype}\PY{p}{)}
\PY{n}{dilate} \PY{o}{=} \PY{n}{cv2}\PY{o}{.}\PY{n}{dilate}\PY{p}{(}\PY{n}{edges}\PY{p}{,} \PY{n}{kernel}\PY{p}{,} \PY{n}{iterations}\PY{o}{=}\PY{l+m+mi}{1}\PY{p}{)}
\PY{c}{\PYZsh{} Find longest line}
\PY{n}{lines} \PY{o}{=} \PY{n}{cv2}\PY{o}{.}\PY{n}{HoughLinesP}\PY{p}{(}\PY{n}{dilate}\PY{p}{,}\PY{l+m+mi}{1}\PY{p}{,}\PY{n}{np}\PY{o}{.}\PY{n}{pi}\PY{o}{/}\PY{l+m+mi}{180}\PY{p}{,}\PY{n}{thHou}\PY{p}{,}
    \PY{n}{minLineLength}\PY{p}{,}\PY{n}{maxLineGap}\PY{p}{)}
\PY{n}{maxL} \PY{o}{=} \PY{l+m+mi}{0}
\PY{k}{if} \PY{n}{lines} \PY{o}{==} \PY{n+nb+bp}{None}\PY{p}{:}
    \PY{n}{a}\PY{p}{,} \PY{n}{b} \PY{o}{=} \PY{n}{np}\PY{o}{.}\PY{n}{nan}\PY{p}{,} \PY{n}{np}\PY{o}{.}\PY{n}{nan}
\PY{k}{else}\PY{p}{:}
    \PY{k}{for} \PY{n}{x1}\PY{p}{,}\PY{n}{y1}\PY{p}{,}\PY{n}{x2}\PY{p}{,}\PY{n}{y2} \PY{o+ow}{in} \PY{n}{lines}\PY{p}{[}\PY{l+m+mi}{0}\PY{p}{]}\PY{p}{:}
        \PY{n}{L} \PY{o}{=} \PY{n}{np}\PY{o}{.}\PY{n}{linalg}\PY{o}{.}\PY{n}{norm}\PY{p}{(}\PY{p}{(}\PY{n}{x1}\PY{o}{\PYZhy{}}\PY{n}{x2}\PY{p}{,}\PY{n}{y1}\PY{o}{\PYZhy{}}\PY{n}{y2}\PY{p}{)}\PY{p}{)}
        \PY{k}{if} \PY{n}{L} \PY{o}{\PYZgt{}} \PY{n}{maxL}\PY{p}{:}
            \PY{n}{maxL} \PY{o}{=} \PY{n}{L}
            \PY{n}{a} \PY{o}{=} \PY{n+nb}{float}\PY{p}{(}\PY{n}{y1} \PY{o}{\PYZhy{}} \PY{n}{y2}\PY{p}{)} \PY{o}{/} \PY{p}{(}\PY{n}{x1} \PY{o}{\PYZhy{}} \PY{n}{x2}\PY{p}{)}
            \PY{n}{b} \PY{o}{=} \PY{n}{y1} \PY{o}{\PYZhy{}} \PY{n}{a} \PY{o}{*} \PY{n}{x1}
\PY{c}{\PYZsh{} a, b being line parameters: y = a * x + b}
\end{Verbatim}

\begin{DUlineblock}{0em}
\item[] 
\end{DUlineblock}
Since we experienced unsatisfactory calibration results (in terms of later reconstruction compounding) at this stage, we passed those through an image compounding algorithm which allows to achieve a good tuning. This is an image based method which uses as input 2 perpendicular sweeps, at approximately 90 degrees, for the same 3D volume \cite{Wein08}. Briefly, a similarity measure (Normalized Cross Correlation, NCC) between the two sweeps was applied to maximize this measure with the final aim to find the calibration parameters relative to the best overlapping between the images. The initial values of this iterative method are the results of the equations-based approach.
A calibration quality assessment was also implemented in terms of precision and accuracy of the calibration parameters obtained. Precision gives an indication of the dispersion of measures around their mean, whereas the accuracy gives an indication of the difference between the mean of the measures and the real value \cite{Hsu06}. For example, this measure can be the known position of a point in space (\emph{Point accuracy}) or the known dimension of an object (\emph{Distance accuracy}).

\subsection{3D voxel array reconstruction%
  \label{d-voxel-array-reconstruction}%
}

The 3D reconstruction is performed by positioning the 2D US scans in the 3D space by using the corresponding pose. The first step is to import the images (DICOM file, standard format for medical imaging) and the synchronized kinematics files (C3D format) containing pose data. A 3D voxel array is then initialized. The 3D voxel array (a parallelepipedon) should be the smallest one containing the sequence of all the repositioned scans, as seen in Figure \DUrole{ref}{voxarrsmall}, in order to avoid RAM waste. To face this issue, in the present package two options are presented: reorienting manually the global reference frame in order to be approximately aligned with the scan direction during the acquisition; on the other hand, by using the Principal Component Analysis (PCA), it is also possible to find the scan direction and thereby realigning the voxel array according to this direction.\begin{figure}[]\noindent\makebox[\columnwidth][c]{\includegraphics[width=\columnwidth]{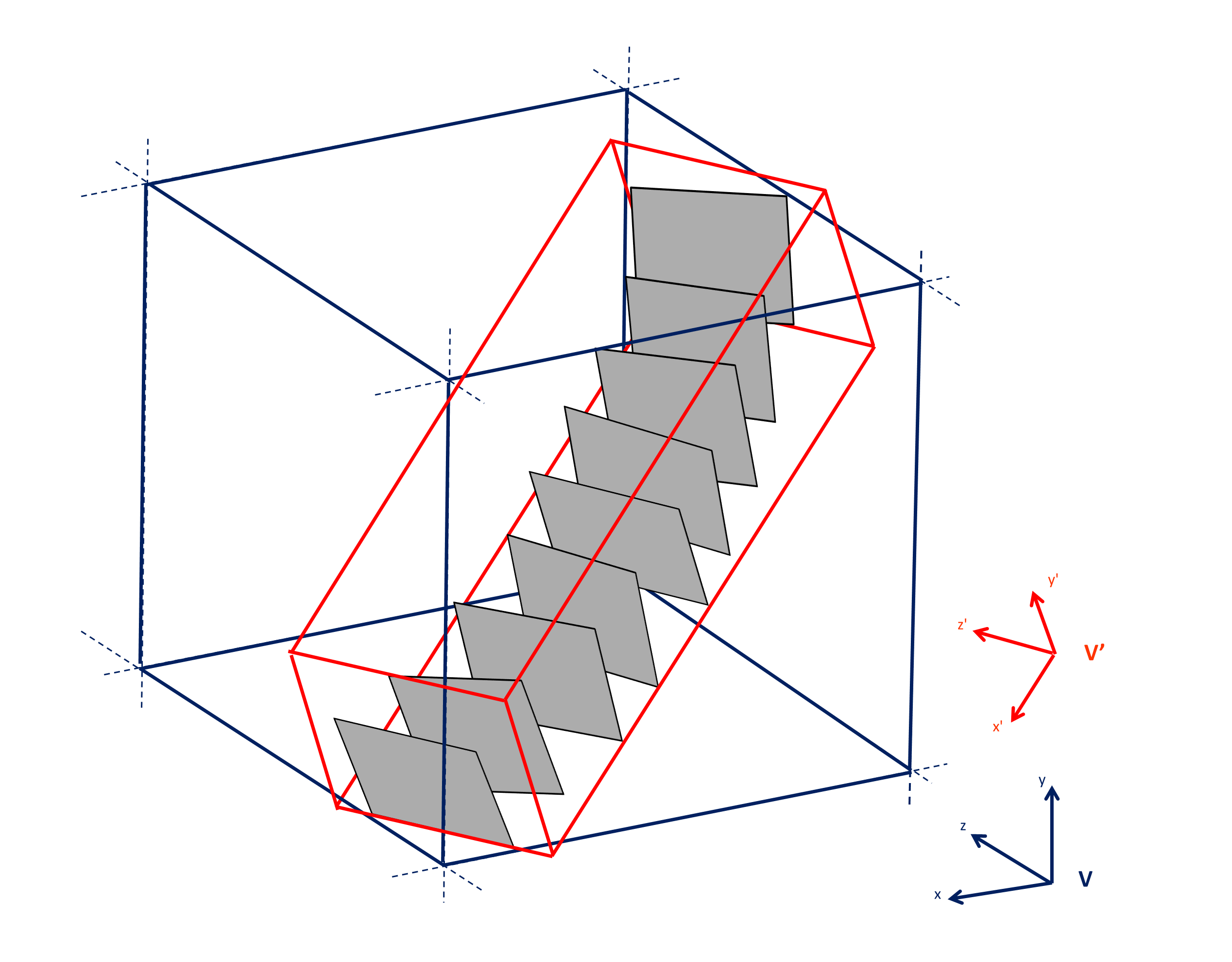}}
\caption{V' is the smallest 3D voxel array parallelepipedon able to contain all the US images. Others can be created, such as V, but they are bigger, occupy more memory and contain more empty voxels. \DUrole{label}{voxarrsmall}}
\end{figure}

The grey values of the original pixels in the 2D slices are then copied in the new corresponding 3D position. This procedure is performed by using an algorithm called Pixel Nearest Neighbor (PNN) which runs through each pixel in every image and fills the nearest voxel with the value of that pixel; in case of multiple contributions to the same voxel, the values are averaged. Below the code to perform this is shown. Each 2D scan is positioned in the 3D volume in a vectorized way.
\begin{DUlineblock}{0em}
\item[] 
\end{DUlineblock}
\begin{Verbatim}[commandchars=\\\{\},fontsize=\footnotesize]
\PY{c}{\PYZsh{} x, y, z: arrays for 3D coordinates of}
\PY{c}{\PYZsh{} the pixels in image I}

\PY{c}{\PYZsh{} idxV: unique ID for each voxel of the}
\PY{c}{\PYZsh{} 3D voxel array}

\PY{c}{\PYZsh{} V: 1D array containing grey values for the}
\PY{c}{\PYZsh{} 3D voxel\PYZhy{}array}

\PY{c}{\PYZsh{} contV: 1D array containing current number of}
\PY{c}{\PYZsh{} contributions for voxels}

\PY{c}{\PYZsh{} I: 2D array containing US slice grey values}

\PY{n}{idxV} \PY{o}{=} \PY{n}{xyz2idx}\PY{p}{(}\PY{n}{x}\PY{p}{,} \PY{n}{y}\PY{p}{,} \PY{n}{z}\PY{p}{,} \PY{n}{xl}\PY{p}{,} \PY{n}{yl}\PY{p}{,} \PY{n}{zl}\PY{p}{)}\PY{o}{.}\PY{n}{astype}\PY{p}{(}\PY{n}{np}\PY{o}{.}\PY{n}{int32}\PY{p}{)}
\PY{n}{V}\PY{p}{[}\PY{n}{idxV}\PY{p}{]} \PY{o}{=} \PY{p}{(}\PY{n}{contV}\PY{p}{[}\PY{n}{idxV}\PY{p}{]} \PY{o}{*} \PY{n}{V}\PY{p}{[}\PY{n}{idxV}\PY{p}{]}\PY{p}{)} \PY{o}{/} \PY{p}{(}\PY{n}{contV}\PY{p}{[}\PY{n}{idxV}\PY{p}{]} \PY{o}{+} \PY{l+m+mi}{1}\PY{p}{)}
    \PY{o}{+} \PY{n}{I}\PY{o}{.}\PY{n}{ravel}\PY{p}{(}\PY{p}{)} \PY{o}{/} \PY{p}{(}\PY{n}{contV}\PY{p}{[}\PY{n}{idxV}\PY{p}{]} \PY{o}{+} \PY{l+m+mi}{1}\PY{p}{)} \PY{c}{\PYZsh{} iterative avg}
\end{Verbatim}

\begin{DUlineblock}{0em}
\item[] 
\end{DUlineblock}
Only 2 outer loops exist, one for the DICOM file number and one for the scan number.
After all the scans are correctly positioned in the 3D space, gaps can occur in the voxel array when the voxel size is small compared to the distance between the acquired images (e.g. scanning velocity significantly different from 0). Therefore interpolation methods are applied for filling these empty voxels. For optimizing this process, a robust method was also used, i.e. convex hull (see Figure \DUrole{ref}{convhull}), for restricting the gap filling operation only to the voxels contained between 2 consecutive slices:\begin{figure}[]\noindent\makebox[\columnwidth][c]{\includegraphics[width=\columnwidth]{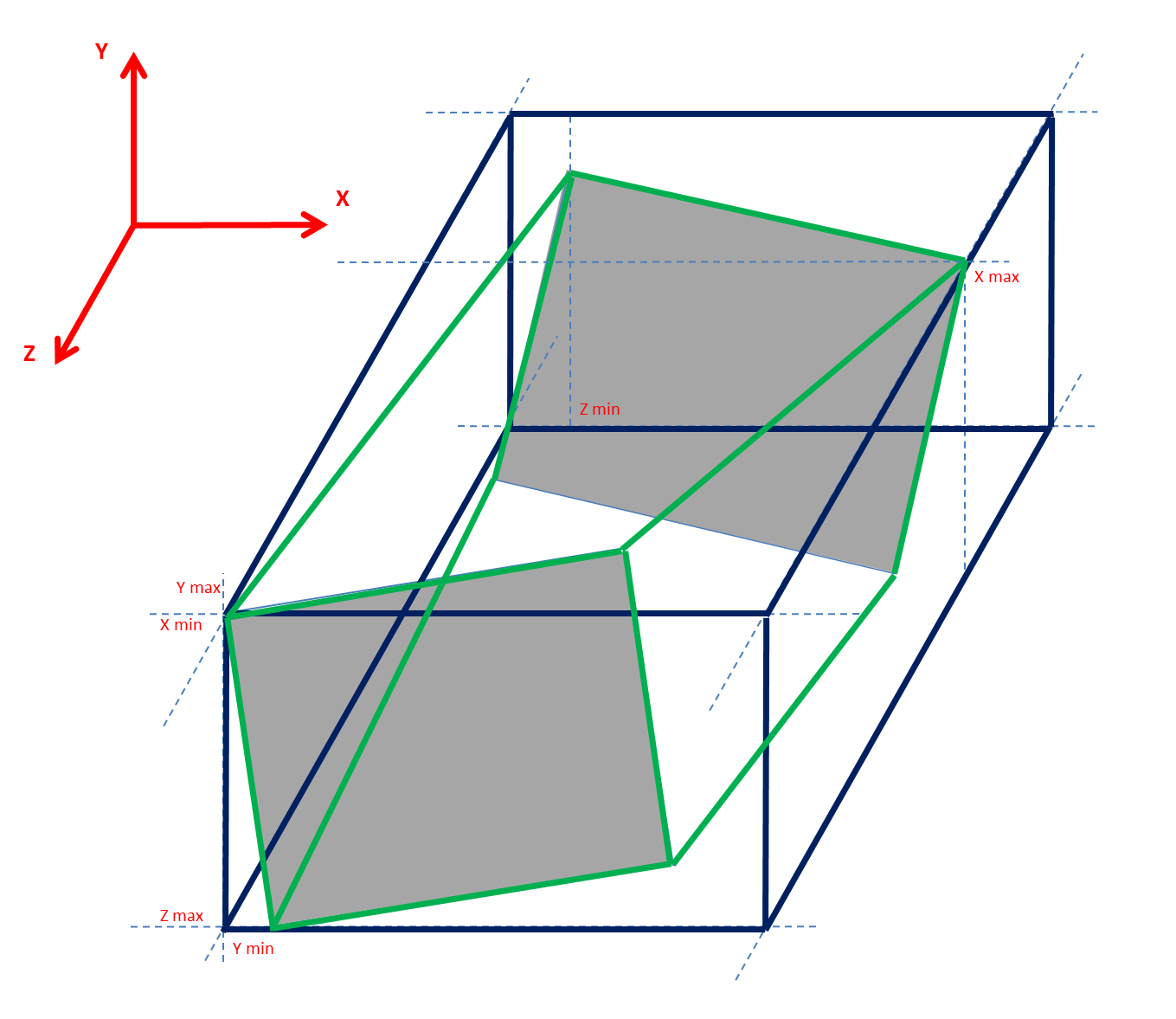}}
\caption{Considering 2 US images consecutive in time, the convex hull is the smaller object able to contain them. An easier shape can be created, such as the parallelepipedon, but this is always bigger in volume. \DUrole{label}{convhull}}
\end{figure}

The quick-and-dirty way, known as VNN (Voxel Nearest Neighbour), consists of filling a gap by using the closest voxel having an assigned grey value. We also implemented another (average cube) solution which consist of the following steps:%
\begin{itemize}

\item 

Create a cube with side 3 voxels, centered around the gap;
\item 

Search the minimum percentage of non-gaps inside the cube (100\% = number of voxels in the cube);
\item 

If that percentage is found, a non-gap voxels average (weighted by the Euclidean distances) is performed into the cube;
\item 

If that percentage is not found, the cube size in incremented by 2 voxels (e.g. 5);
\item 

If cube size is lesser or equal than a maximum size, start again from point 2. Otherwise, stop and don't fill the gap.
\end{itemize}

The entire voxel array can be subdivided in N parallelepipedal blocks, and the gap filling is performed on each one at a time, to spare some of the RAM. The bigger the number of blocks, the bigger the number of iterations to go, but the smaller the block size, the RAM used and the time spent per iteration.
Finally, both the voxel array scans silhouette (previously created with the wrapping convex hulls) and the grey scale data voxel array are exported to VTI files, after being converted to \texttt{vtk.vtkImageData}. These can be opened with software like MeVisLab or Paraview for visualization and further processing.

\subsection{Preliminary results%
  \label{preliminary-results}%
}

The calibration quality assessments were 1.9 mm and 3.9 mm for the distance accuracy and reconstruction precision, respectively. The average data processing time (calibration + reconstruction + gap filling) over 3 trials on a human calf, shown in Figure \DUrole{ref}{calf}, was 5.9 min, on a 16 GB RAM Intel i7 2.7 GHz machine.\begin{figure}[]\noindent\makebox[\columnwidth][c]{\includegraphics[width=\columnwidth]{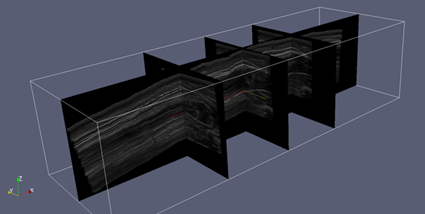}}
\caption{Three transversal and one longitudinal section of a reconstructed 3D voxel array (human calf scanning, about 90M voxels, $10mm^3$ each). \DUrole{label}{calf}}
\end{figure}

\end{document}